\documentclass[11pt,a4paper]{article}
\usepackage[hyperref]{acl2020}
\usepackage{times}
\usepackage{latexsym}

\usepackage{microtype}

\aclfinalcopy 

\usepackage{multirow}
\usepackage{amsmath}
\usepackage{ifthen}
\usepackage{amssymb}
\usepackage{graphicx}
\usepackage{tabu}
\usepackage{booktabs}
\usepackage{subcaption}
\usepackage{url}

\title{Improving Non-autoregressive Neural Machine Translation with Monolingual Data}

\author{Jiawei Zhou \\
  Harvard University \\ 
  \texttt{jzhou02@g.harvard.edu} \\\And
  Phillip Keung \\
  Amazon Inc. \\ 
  \texttt{keung@amazon.com} \\}

\date{}

\begin{document}
\maketitle
\begin{abstract}

Non-autoregressive (NAR) neural machine translation is usually done via knowledge distillation from an autoregressive (AR) model. Under this framework, we leverage large monolingual corpora to improve the NAR model's performance, with the goal of transferring the AR model's generalization ability while preventing overfitting. On top of a strong NAR baseline, our experimental results on the WMT14 En-De and WMT16 En-Ro news translation tasks confirm that monolingual data augmentation consistently improves the performance of the NAR model to approach the teacher AR model's performance, yields comparable or better results than the best non-iterative NAR methods in the literature and helps reduce overfitting in the training process.


\end{abstract}

\section{Introduction}

Neural machine translation (NMT) \citep{sutskever2014sequence, bahdanau2014neural} has achieved impressive performance in recent years, but the autoregressive decoding process limits the translation speed and restricts low-latency applications. To mitigate this issue, many non-autoregressive (NAR) translation methods have been proposed, including latent space models \citep{gu2017non, ma2019flowseq, shu2019latent}, iterative refinement methods \citep{lee2018deterministic, ghazvininejad2019mask}, and alternative loss functions \citep{libovicky2018end, wang2019non, wei2019imitation, li2019hint, shao2019minimizing}.
The decoding speedup for NAR models is typically 2-15$\times$ depending on the specific setup (e.g., the number of length candidates, number of latent samples, etc.), and NAR models can be tuned to achieve different trade-offs between time complexity and decoding quality \citep{gu2017non, wei2019imitation, ghazvininejad2019mask, ma2019flowseq}.

Although different in various aspects, all of these methods are based on transformer modules \citep{vaswani2017attention}, and depend on a well-trained AR model to obtain its output translations to create targets for NAR model training. This training setup is well-suited to leverage external monolingual data, since the target side of the NAR training corpus is always generated by an AR model. Techniques like backtranslation \citep{sennrich2015improving} are known to improve MT performance using monolingual data alone. However, to the best of our knowledge, monolingual data augmentation for NAR-MT has not been reported in the literature.

In typical NAR-MT model training, an AR teacher provides a consistent supervision signal for the NAR model; the source text that was used to train the teacher is decoded by the teacher to create synthetic target text. In this work, we use a large amount of source text from monolingual corpora to generate additional teacher outputs for NAR-MT training.

We use a transformer model with minor structural changes to perform NAR generation in a non-iterative way, which establishes stronger baselines than most of the previous methods. We demonstrate that  generating additional training data with monolingual corpora consistently improves the translation quality of our baseline NAR system on the WMT14 En-De and WMT16 En-Ro translation tasks. Furthermore, our experiments show that NAR models trained with increasing amount of extra monolingual data are less prone to overfitting and generalize better on longer sentences.

In addition, we have obtained Ro$\rightarrow$En and En$\rightarrow$De results which are state-of-the-art for non-iterative NAR-MT, just by using more monolingual data.




\section{Methodology}

\subsection{Basic Approach}

Most of the previous methods treat the NAR modeling objective as a product of independent token probabilities \citep{gu2017non}, but we adopt a different point of view by simply treating the NAR model as a function approximator of an existing AR model.

Given an AR model and a source sentence, the translation process of the greedy output\footnote{By `greedy', we mean decoding with a beam width of 1.} of the AR model is a complex but deterministic function. Since the neural networks can be near-perfect non-linear function approximators \citep{liang2016deep}, we can expect an NAR model to learn the AR translation process quite well, as long as the model has enough capacity. In particular, we first obtain the greedy output of a trained AR model, and use the resulting paired data to train the NAR model. Other papers on NAR-MT \citep{gu2017non, lee2018deterministic, ghazvininejad2019mask} have used AR teacher models to generate training data, and this is a form of sequence-level knowledge distillation \citep{kim2016sequence}.



\begin{table}
\centering
\begin{tabular}{@{}lccc@{}}  
\toprule
 & Parallel & En Mono.  & Non-En Mono. \\ 
\midrule
En-Ro & 608,320  & 2,197,792  & 2,261,206 \\
En-De & 4,459,186  & 3,008,621  & 3,015,110 \\
\bottomrule
\end{tabular}
\caption{Number of sentences per language arc. `Mono' refers to the amount of monolingual text available.}
\label{table:data}
\end{table}

\subsection{Model Structure}
\label{ssec:structure}
Throughout this paper, we focus on non-iterative NAR methods. We use standard transformer structures with a few small changes for NAR-MT, which we describe below.



For the target side input, most of the previous work simply copied the source side as the decoder's input. We propose a \emph{soft copying method} by using a Gaussian kernel to smooth the encoded source sentence embeddings $x^{enc}$. Suppose the source and target lengths are $T$ and $T'$ respectively. Then the $t$-th input token for the decoder is $\sum_{i=1}^{T} x^{enc}_i\cdot K(i, t)$, where $K(i, t)$ is the Gaussian distribution evaluated at $i$ with mean $\frac{T}{T'}t$ and variance $\sigma^2$. ($\sigma^2$ is a learned parameter.)

We modify the decoder self-attention mask so that it does not mask out the future tokens, and every token is dependent on both its preceding and succeeding tokens in every layer.

\citet{gu2017non}, \citet{lee2018deterministic}, \citet{li2019hint} and \citet{wang2019non} use an additional positional self-attention module in each of the decoder layers, but we do not apply such a layer. It did not provide a clear performance improvement in our experiments, and we wanted to reduce the number of deviations from the base transformer structure. Instead, we add positional embeddings at each decoder layer.

\subsection{Length Prediction}
\label{ssec:length_pred}
We use a simple method to select the target length for NAR generation at test time \citep{wang2019non,li2019hint}, where we set the target length to be $T' = T + C$, where $C$ is a constant term estimated from the parallel data and $T$ is the length of the source sentence. We then create a list of candidate target lengths ranging from $[T'-B, T'+B]$ where $B$ is the half-width of the interval. For example, if $T=5$, $C=1$ and we used a half-width of $B=2$, then we would generate NAR translations of length $[4,5,6,7,8]$, for a total of 5 candidates. These translation candidates would then be ranked by the AR teacher to select the one with the highest probability. This is referred to as length-parallel decoding in \citet{wei2019imitation}.

\section{NAR-MT with Monolingual Data}

Augmenting the NAR training corpus with monolingual data provides some potential benefits. Firstly, we allow more data to be translated by the AR teacher, so the NAR model can see more of the AR translation outputs than in the original training data, which helps the NAR model generalize better. Secondly, there is much more monolingual data than parallel data, especially for low-resource languages.

Incorporating monolingual data for NAR-MT is straightforward in our setup. Given an AR model that we want to approximate, we obtain the source-side monolingual text and use the AR model to generate the targets that we can train our NAR model on.

\begin{table*}[t!]
\begin{center}
\begin{tabular}{@{}lcccc@{}}  
\toprule
\multirow{2}{*}{Models} & \multicolumn{2}{c}{WMT16} & \multicolumn{2}{c}{\, WMT14} \\
 & En$\rightarrow$Ro  & Ro$\rightarrow$En & En$\rightarrow$De  & De$\rightarrow$En \\ 
\midrule
NAT-FT \cite{gu2017non} & 27.29 & 29.06 & 17.69 & 21.47 \\
NAT-FT (+NPD s=10) & 29.02 & 30.76 & 18.66 & 22.41 \\
NAT-FT (+NPD s=100) & 29.79 & 31.44 & 19.17 & 23.20 \\
NAT-IR ($i_{dec}$=1) \cite{lee2018deterministic} & 24.45 & 25.73 & 13.91 & 16.77 \\
CTC \cite{libovicky2018end} & 19.93 & 24.71 & 17.68 & 19.80 \\
imitate-NAT \cite{wei2019imitation} &  28.61 & 28.90 &  22.44 & 25.67 \\
imitate-NAT (+LPD) & 31.45 & 31.81 & 24.15 & 27.28 \\
CMLM \cite{ghazvininejad2019mask} & 27.32 & 28.20 & 18.05 & 21.83 \\
FlowSeq \cite{ma2019flowseq} &  29.73 & 30.72 &  23.72 & 28.39 \\
FlowSeq (NPD n=30) &  \textbf{32.20} & \textbf{32.84} & \textbf{25.31}  & \textbf{30.68} \\
\midrule
Our AR Transformer (beam 1) & 33.56 & 33.68 & 28.84 & 32.77 \\
Our AR Transformer (beam 4) & 34.50 & 34.01 & 29.65 & 33.65 \\
\midrule
Our NAR baseline ($B$=5) & 31.21 & 32.06 & 23.57 & 29.01  \\
+ 50\% monolingual data & 31.74 & 33.16 & 25.35 & 29.78 \\
+ monolingual data & 31.91 & 33.46 & 25.53 & 29.96 \\
+ monolingual data and de-dup & \textbf{31.96} & \textbf{33.57} & \textbf{25.73} & \textbf{30.18} \\
+ monolingual data and de-dup (doubled training time)$^\dag$ & \textbf{32.30} & \textbf{33.56} & \textbf{26.54} & \textbf{30.80} \\
\bottomrule
\end{tabular}
\end{center}
\caption{BLEU scores on the WMT16 En-Ro and WMT14 En-De test sets for different NAR models. All reported scores are from non-iterative NAR methods with similar hyper-parameter settings for transformers. `de-dup' removes adjacent duplicated tokens. $B$ is the half-width in Sec. \ref{ssec:length_pred}. $^\dag$ marks results obtained by simply training the model for longer time until full convergence to fairly compare with previous works, with which we achieve further SOTA performance, but they are not directly comparable to our other experiments and are thus ignored in our discussion and analysis.}
\label{table:results}
\end{table*}

\section{Experimental Setup}

\paragraph{Data}
We evaluate NAR-MT training on both the WMT16 En-Ro (around 610k sentence pairs) and the WMT14 En-De (around 4.5M sentence pairs) parallel corpora along with the associated WMT monolingual corpora for each language. For the parallel data, we use the processed data from \citet{lee2018deterministic} to be consistent with previous publications. The WMT16 En-Ro task uses newsdev-2016 and newstest-2016 as development and test sets, and the WMT14 En-De task uses newstest-2013 and newstest-2014 as development and test sets.
We report all results on test sets.
We used the Romanian portion of the News Crawl 2015 corpus and the English portion of the Europarl v7/v8 corpus\footnote{http://www.statmt.org/wmt16/translation-task.html\label{mono-url}} as monolingual text for our En-Ro experiments, which are both about 4 times larger than the original paired data.
We used the News Crawl 2007/2008 corpora for German and English monolingual text\textsuperscript{\ref{mono-url}} in our En-De experiments, and downsampled them to ${\sim}3$ million sentences per language.
The data statistics are summarized in Table \ref{table:data}.
The monolingual data are processed following \citet{lee2018deterministic}, which are tokenized and segmented into subword units \citep{sennrich2015neural}. The vocabulary is shared between source and target languages and has ${\sim}40$k units. We use BLEU to evaluate the translation quality\footnote{We report tokenized BLEU scores in line with prior work \citep{lee2018deterministic, ma2019flowseq}, which are case-insensitive for WMT16 En-Ro and case-sensitive for WMT14 En-De in the data provided by \citet{lee2018deterministic}.}.

\paragraph{Model Configuration}
We use the settings for the base transformer configuration in \citet{vaswani2017attention} for all the models: 6 layers per stack, 8 attention heads per layer, 512 model dimensions and 2048 hidden dimensions. 
The AR and NAR model have the same encoder-decoder structure, except for the decoder attention mask and the decoding input for the NAR model as described in Sec. \ref{ssec:structure}.

\paragraph{Training and Inference}
We initialize the NAR embedding layer and encoder parameters with the AR model's. The NAR model is trained with the AR model's greedy outputs as targets. We use the Adam optimizer, with batches of size 64k tokens for one gradient update, and the learning rate schedule is the same as the one in \citet{vaswani2017attention}, where we use 4,000 warm-up steps and the maximum learning rate is around 0.0014. We stop training when there is no further improvement in the last 5 epochs, and training finishes in 30 epochs for AR models and 50 epochs for NAR models, except for the En-De experiments with monolingual data where we train for 35 epochs to roughly match the number of parameter updating steps without using extra monolingual data (${\sim}140$k steps). We average the last 5 checkpoints to obtain the final model. We train the NAR model with cross-entropy loss and label smoothing ($\epsilon=0.1$). During inference time, we use length parallel decoding with $C=0$, and evaluate the BLEU scores on the reference sentences.
All the models are implemented with MXNet and GluonNLP \citep{gluoncvnlp2019}.
We used 4 NVIDIA V100 GPUs for training, which takes about a day for an AR model and up to a week for an NAR model depending on the data size, and testing is performed on a single GPU.

\section{Results and Analysis}

\begin{figure}
    \centering
    \includegraphics[width=0.5\textwidth]{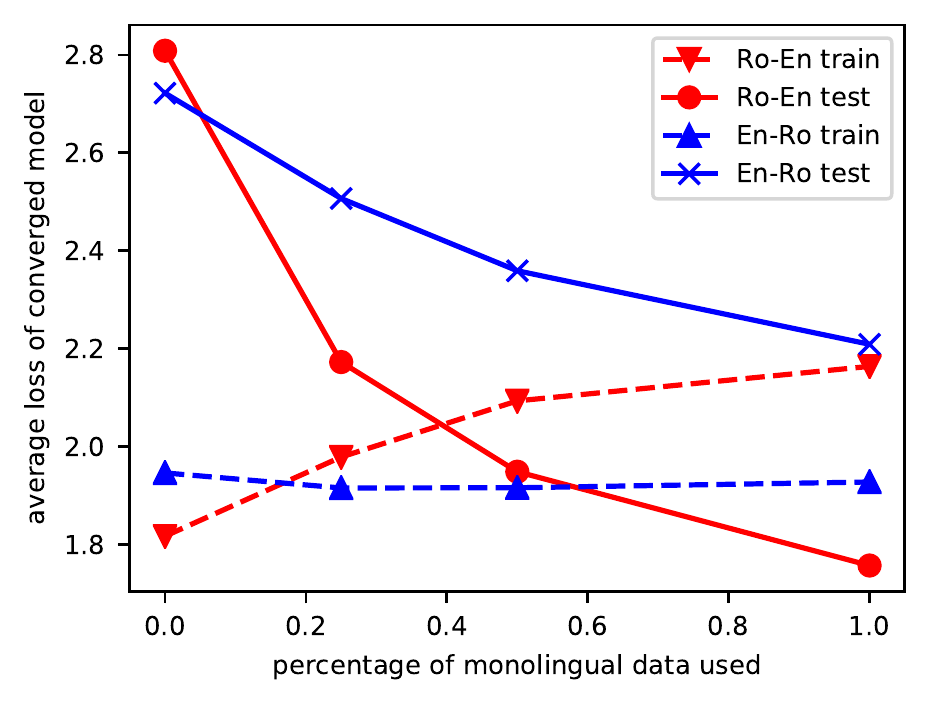}
    \caption{Average loss of the NAR models versus the percentage of monolingual data used during training. The test set losses decrease as more monolingual data is added, and the gap towards training losses are closing, which indicates that monolingual data augmentation reduces overfitting.}
    \label{fig:loss}
\end{figure}

\paragraph{Main Results} We present our BLEU scores alongside the scores of other non-iterative methods in Table \ref{table:results}. Our baseline results surpass many of the previous results, which we attribute to the way that we initialize the decoding process. Instead of directly copying the source embeddings to the decoder input, we use an interpolated version of the encoder outputs as the decoder input, which allows the encoder to transform the source embeddings into a more usable form. Note that a similar technique is adopted in \citet{wei2019imitation}, but our model structure and optimization are much simpler as we do not have any imitation module for detailed teacher guidance. 

Our results confirm that the use of monolingual data improves the NAR model's performance,
and the gain is proportional
to the amount of extra monolingual data as seen
from the BLUE scores with using only half and the
full monolingual data.
By incorporating all of the monolingual data for the En-Ro NAR-MT task, we see a gain of 0.70 BLEU points for the En$\rightarrow$Ro direction and 1.40 for the Ro$\rightarrow$En direction. Similarly, we also see significant gains in the En-De NAR-MT task, with an increase of 1.96 BLEU points for the En$\rightarrow$De direction and 0.95 for the De$\rightarrow$En direction.

By removing the duplicated output tokens as a simple postprocessing step (following \citet{lee2018deterministic}), we achieved 33.57 BLEU for the WMT16 Ro$\rightarrow$En direction and 25.73 BLEU for the WMT14 En$\rightarrow$De direction, which are state-of-the-art among non-iterative NAR-MT results. In addition, our work shrinks the gap between the AR teacher and the NAR model to just 0.11 BLEU points in the Ro$\rightarrow$En direction.


\begin{table}[t!]
\begin{center}
\resizebox{\columnwidth}{!}{%
\begin{tabular}{@{}c ccc ccc@{}}  
\toprule
 & \multicolumn{3}{c}{En$\rightarrow$Ro}  & \multicolumn{3}{c}{Ro$\rightarrow$En} \\ 
 \cmidrule(r){2-7}
 & no & half & all & no & half & all \\
B & mono & mono & mono & mono & mono & mono \\
\midrule
0 & 27.19 & +0.65 & +0.56 & 26.62 & +1.52 & +1.58 \\
1 & 29.34 & +0.63 & +0.69 & 28.81 & +1.26 & +1.46 \\
2 & 30.46 & +0.34 & +0.45 & 30.18 & +1.08 & +1.24 \\
3 & 30.87 & +0.37 & +0.71 & 31.24 & +0.88 & +1.09 \\
4 & 31.06 & +0.45 & +0.67 & 31.92 & +0.90 & +1.25 \\
5 & 31.21 & +0.53 & +0.70 & 32.06 & +1.10 & +1.40 \\
6 & 31.20 & +0.39 & +0.62 & 31.98 & +1.17 & +1.43 \\
7 & 30.99 & +0.43 & +0.51 & 31.85 & +1.19 & +1.31 \\
\midrule
gold & 29.64 & +0.61 & +0.85 & 29.83 & +1.42 & +1.69 \\
\bottomrule
\end{tabular}%
}
\end{center}
\caption{BLEU scores on the WMT16 En-Ro test sets for NAR models trained with different numbers of length candidates and amounts of additional monolingual data.  The half-width B determines the number of length candidates (Sec. \ref{ssec:length_pred}). `gold' refers to using the true target length instead of predicting it. All the +deltas are relative to the `no mono' case.}
\label{table:length}
\end{table}

\paragraph{Losses in Training and Evaluation}
To further investigate how much the monolingual data contributes to BLEU improvements, we train En-Ro NAR models with 0\%, 25\%, 50\%, and 100\% of the monolingual corpora and plot the cross-entropy loss on the training data and the testing data for the converged model. In Figure~\ref{fig:loss}, when no monolingual data is used, the training loss typically converges to a lower point compared to the loss on the testing set, which is not the case for the AR model where the validation and testing losses are usually lower than the training loss. This indicates that the NAR model overfits to the training data, which hinders its generalization ability. However, as more monolingual data is added to the training recipe, the overfitting problem is reduced and the gap between the evaluation and training losses shrinks. 


\paragraph{Effect of Length-Parallel Decoding}
To test how the NAR model performance and the monolingual gains are affected by the number of decoding length candidates, we vary the half-width $B$ (Sec. \ref{ssec:length_pred}) across a range of values and test the NAR models trained with 0\%, 50\%, and 100\% of the monolingual data for the En-Ro task (Table \ref{table:length}). The table shows that having multiple length candidates can increase the BLEU score significantly and can be better than using the gold target length, but having too many length candidates can hurt the performance and slow down decoding (in our case, the optimal $B$ is 5). Nonetheless, for every value of $B$, the BLEU score consistently increases when monolingual data is used, and more data brings greater gains.

\paragraph{BLEU under Different Sentence Lengths}
In Table \ref{table:bleuwithlength}, we present the BLEU scores on WMT16 Ro$\rightarrow$En test sentences grouped by source sentence lengths. We can see that the baseline NAR model's performance drops quickly as sentence length increases, whereas the NAR model trained with monolingual data degrades less over longer sentences, which demonstrates that external monolingual data improves the NAR model's generalization ability.


\begin{table}[t!]
\begin{center}
\resizebox{\columnwidth}{!}{%
\begin{tabular}{@{}cccccc@{}}  
\toprule
src    & \#    & AR     & NAR  & +half & +all \\ 
length & sent. & beam 1 & baseline & mono & mono \\
\midrule
$[1, 20]$    & 865  & 32.12 & 29.96 & 30.94 & 31.10 \\
$[21, 40]$   & 867  & 33.82 & 30.77 & 31.92 & 31.96 \\
$[41, 60]$   & 228  & 35.13 & 29.59 & 31.33 & 31.81 \\
$[61, 80]$   & 29   & 35.09 & 26.69 & 27.99 & 30.47 \\
$[81, 120]$  & 8    & 34.13 & 16.47 & 28.92 & 29.47 \\
$[121, 140]$ & 2    & 6.70  & 3.11  & 3.56  & 5.99  \\
\bottomrule
\end{tabular}%
}
\end{center}
\caption{BLEU scores for source sentences in different length intervals on the WMT16 Ro$\rightarrow$En test set. The gold target length is provided during decoding.}
\label{table:bleuwithlength}
\end{table}

\section{Discussion}

We found that monolingual data augmentation reduces overfitting and improves the translation quality of NAR-MT models. We note that the monolingual corpora are derived from domains which may be different from those of the parallel training data or evaluation sets, and a mismatch can affect NAR translation performance. Other work in NMT has examined this issue in the context of backtranslation (e.g., \citet{edunov2018understanding}), and we expect the conclusions to be similar in the NAR-MT case.

There are several open questions to investigate: Are the benefits of monolingual data orthogonal to other techniques like iterative refinement? Can the NAR model perfectly recover the AR model's performance with much larger monolingual datasets? Are the observed improvements language-dependent? We will consider these research directions in future work.

\bibliography{acl2020}
\bibliographystyle{acl_natbib}

\end{document}